%% file: latex/acl_latex.tex
\pdfoutput=1
\documentclass[11pt]{article}

\usepackage[final]{latex/acl}

\usepackage{times}
\usepackage{latexsym}
\usepackage{amssymb}
\usepackage{amsmath}
\usepackage{booktabs}
\usepackage{multirow}

\usepackage[T1]{fontenc}
\usepackage[utf8]{inputenc}

\usepackage{microtype}

\usepackage{inconsolata}

\usepackage{graphicx}

\title{Generalizing to Unseen Disaster Events: A Causal View}

\author{
 \textbf{Philipp Seeberger}, 
 \textbf{Steffen Freisinger}, 
 \textbf{Tobias Bocklet},
 \textbf{Korbinian Riedhammer}
\\
 Technische Hochschule Nürnberg Georg Simon Ohm \\
  \small$\texttt{\{philipp.seeberger,steffen.freisinger,tobias.bocklet,korbinian.riedhammer\}@th-nuernberg.de}$
}

\begin{document}

\maketitle
\begin{abstract}
Due to the rapid growth of social media platforms, these tools have become essential for monitoring information during ongoing disaster events. 
However, extracting valuable insights requires real-time processing of vast amounts of data. 
A major challenge in existing systems is their exposure to event-related biases, which negatively affects their ability to generalize to emerging events. 
While recent advancements in debiasing and causal learning offer promising solutions, they remain underexplored in the disaster event domain. 
In this work, we approach bias mitigation through a causal lens and propose a method to reduce event- and domain-related biases, enhancing generalization to future events. 
Our approach outperforms multiple baselines by up to +1.9\% F1 and significantly improves a PLM-based classifier across three disaster classification tasks.
\end{abstract}

\input{latex/1_introduction}
\input{latex/2_method}
\input{latex/3_experiments}
\input{latex/4_conclusion}
\input{latex/5_limitations}
\input{latex/6_ethics}

% \section*{Acknowledgments}
% TODO

\bibliography{latex/custom.bib}

\appendix

\section{Appendix}
\label{sec:appendix}
\input{latex/7_appendix}

\end{document}

%% file: latex/1_introduction.tex
\section{Introduction}\label{sec:introduction}

Social media has become a crucial source of information during both natural (e.g., hurricanes) and human-made disasters (e.g., bombings) \citep{reuter2018crisis}. 
Unlike traditional news sources, social media provides real-time updates, firsthand observations, and insights via affected individuals \citep{sakaki2010earthquake}.
Filtering these information nuggets is essential for situational awareness and for supporting relief organizations, government agencies, and emergency responders \citep{kruspe2021review}.

A major challenge lies in processing the vast volume of social media data, requiring automated methods to reliably detect relevant content \citep{kaufhold2021crisis}.
Although recent advances in Large Language Models (LLMs) demonstrate promising capabilities, their considerably higher latency limits their applicability for this task (see \autoref{table:experiments:llms}), making smaller Pretrained Language Models (PLMs) a necessary alternative. 
Recent research has explored binary, multi-class, and multi-label classification to categorize posts into broad (e.g., \textit{Relevant} vs. \textit{Irrelevant}) or fine-grained (e.g., \textit{Infrastructure Damage}, \textit{Missing People}, etc) categories \citep{olteanu2015crisislex,alam2021humaid,buntain2021trecis}.

Another key challenge is the scarcity and absence of in-domain data. 
Emerging disasters are unpredictable and past event data often fails to generalize due to shifts in event-specific (e.g., locations) and domain-specific (e.g., wildfire spread patterns) features \citep{medina2020eann}. 
Additionally, social media posts are typically short, noisy, and lack contextual depth, making it difficult for models to adapt to unseen disaster events \citep{wiegmann2020analysis}.

To mitigate biased models, prior work has investigated domain adaptation \citep{alam2018domainadaptation,seeberger2022emlm} and adversarial learning methods \citep{medina2020eann}, but these approaches struggle with mixed event types and rely on large amounts of data. 
Other debiasing techniques have been extensively studied in related areas such as fake news detection \citep{zhu2022endef}, sentiment analysis \citep{chew2024nfl}, and question answering \citep{clark2019poe}, but have never been applied to the disaster response domain.

Recently, causal learning has gained attention for debiasing by modeling cause-effect relationships \citep{wei2021macr,qian2021corsair,zhu2022endef,chen2023causal,zhan2024evolving}.
However, the causal perspective remains underexplored for disaster event modeling.
In this work, we adopt a causal view and propose a method to mitigate event- and domain-related\footnote{We consider event types (e.g., hurricanes) as domains.} biases, improving generalization to future disaster events.

\paragraph{Contributions} (1) We present a causal perspective on event- and domain-related biases in real-time disaster classification and propose a framework for improved generalization.
(2) We reproduce and adapt a broad range of debiasing methods, demonstrating the effectiveness of our approach on three real-world disaster classification datasets.

%% file: latex/2_method.tex
\section{Method}\label{sec:method}

\begin{figure}[t]
  \centering
\includegraphics[width=1.0\linewidth]{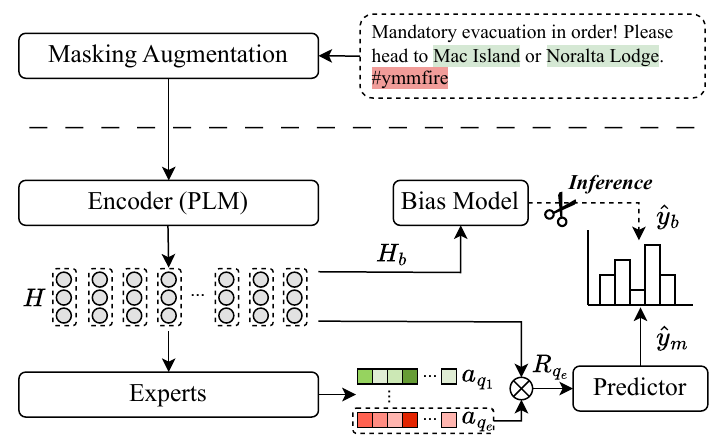}
  \caption{Overview of the proposed framework. The masking augmentation and bias model are used only during training. During inference, the bias model is removed to obtain debiased predictions. The experts and predictor corresponds to the main model, and only a single expert’s output $R_q$ is used for final prediction.}
  \label{fig:method:framework}
\end{figure}

Let $X=\{(p_i,y_i)\}_{i=1}^{N}$ denote a collection of social media posts, where each post $p = (w_1,\dots,w_n)$ is a sequence of $n$ tokens with assigned ground truth label $y \in \{1,\dots,l\}$ indicating one of $l$ information types.
The goal is to learn a classifier that predicts the correct information type for new posts.
Therefore, each post is encoded by a PLM encoder into a sequence of contextualized representations $H = (h_1,\dots,h_m) \in \mathbb{R}^{m \times d}$ with encoder sequence length $m$ and hidden dimension $d$.
The resulting representations are aggregated and passed through a classification layer to predict the information type $\hat y$.
However, as shown in prior work \cite{medina2020eann}, models trained on disaster-related posts often rely on spurious event-specific cues (e.g., locations, hashtags) and domain-related patterns (e.g., hurricanes, bombings), which hinder generalization to unseen events.
To address this issue, our method explicitly disentangles biased signals from more generalizable signals, thereby improving model robustness and transferability across diverse disaster scenarios.
An overview of our proposed framework is shown in \autoref{fig:method:framework}.

\subsection{Event-related Bias}\label{sec:method:eventbias}

To remove spurious event-specific correlations, we follow the causal frameworks of \citet{wei2021macr,zhu2022endef} and model event-related bias using a causal graph with a direct effect path $E \to Y$ and indirect effect path $E \to P \to Y$, where $E$, $P$, and $Y$ represent event context, post, and information type, respectively.
For example, the event \textit{Jakarta floods} causes certain tokens to appear in the post, such as \textit{\#JakartaFlood}, which may introduce spurious shortcuts w.r.t. the information type.
To mitigate such confounding effects, our goal is to block the direct path $E \to Y$ while preserving disaster-related features captured through $P$.
We achieve this by identifying event-specific tokens and model their direct contribution to the predicted information type.
During inference, we remove the estimated direct effect to obtain debiased predictions.

\paragraph{Identification} 
First, we identify the observable bias tokens $u=(u_1,u_2,\dots)$ for each $p$ as proxies for the event context.
Specifically, we extract named entities (e.g., persons, locations, buildings), Twitter-specific markers such as hashtags (i.e., retrieval keywords) and numerical values (e.g., casualties).
These tokens introduce potential event-specific bias that must be considered during model training and inference.

\paragraph{Modeling}
Next, we explicitly model the direct effect $E \to Y$ using a bias model consisting of a bias encoder and predictor. 
The model receives the counterfactual contextualized representations $H_{b}$ of $p_{b} = (\text{[CLS]}, u_1, \text{[SEP]}, u_2, \text{[SEP]}, \dots)$ as input and produces counterfactual predictions $\hat y_b$.
We optimize the model with cross-entropy loss:
\begin{equation}
    \mathcal{L}_{bias} = - \sum_{(p,y) \in X} y \, log(\hat y_b)
\end{equation}
This enables the bias model to capture the direct effect of event context on the main task, which is later integrated into the main model and inference process.

\subsection{Domain-related Bias}\label{sec:method:domainbias}

When training on data with mixed event types, domain-related bias arises as overrepresented event types dominate the model’s attention patterns and decision boundaries, resulting in degraded cross-domain robustness \cite{medina2020eann}.
However, directly blocking these causal paths is challenging.
Inspired by \citet{wu2024das}, we propose a query-based approach that leverages domain-specific experts $Q = \{q_1, \dots, q_e\}$, each corresponding to an event type.
These queries encode domain-specific priors that guide how attention aggregates information from the contextualized representations.
While a single shared query would bias attention toward frequent domains, domain-specific experts encourage balanced representations and mitigate overrepresentation bias.
Unlike \citet{wu2024das}, which employ predefined label-based queries, our method introduces domain-aware experts to reduce interference across event types.

\paragraph{Modeling}
We implement the main model using an attention-based classifier that generates $e$ domain-specific attention distributions $\{a_{q_{1}}, \dots, a_{q_{e}}\}$, where each $a \in \mathbb{R}^{m \times 1}$ corresponds to the attention weights for an expert.
The attention distribution for the assigned expert $q$ is computed as follows: $a_{q} = \text{softmax}(H W_{q} + b_{q})$, where $W_{q} \in \mathbb{R}^{d \times 1}$ and $b_{q} \in \mathbb{R}$ are learnable parameters.
Next, we obtain the final representation for an expert as $R_{q} = \sum_{i} a_{q}^{(i)} h_{i}$ and get the model predictions $\hat y_m$ via a shared predictor.
Both the PLM encoder and the predictor are shared across event types, while only the attention queries are domain-specific.
The rationale for this design is to condition the attention mechanism on the event type, while maintaining knowledge transfer.

The final prediction is fused as $\hat y =   (1 - \alpha) \hat y_m + \alpha \hat y_b$ and optimized with the cross-entropy loss:
\begin{equation}
    \mathcal{L}_{main} = - \sum_{(p,y) \in X} y \, log(\hat y)
\end{equation}
Here, the parameter $\alpha$ controls the contribution of the bias model.
This allows the main model to focus on disaster-relevant features while minimizing the influence of overrepresented event types.

\subsection{Masking Augmentation}\label{sec:method:masking}

To further improve the generalization ability of the models, we introduce \textit{Masking Augmentation}.
For each post, we re-use the identified bias tokens $u=(u_1,u_2,\dots)$ (see \ref{sec:method:eventbias}) and randomly mask each token $u_i$ by replacing it with a special token $\text{[MASK]}$ during training.
From a causal view, masking acts as intervention which generates counterfactual versions of the post, helping to break spurious correlations between individual tokens and target labels.
This encourages the model to rely more on contextual information rather than spurious tokens.

\subsection{Training and Inference}\label{sec:method:inference}

The bias and main model are jointly trained by optimizing the combined loss:
\begin{equation}
    \mathcal{L} = \mathcal{L}_{main} + \lambda \, \mathcal{L}_{bias},
\end{equation}
where $\lambda$ is a trade-off parameter that controls the strength of bias mitigation.
Note that the parameters of each expert are updated only for samples with the corresponding event type.
We also explored to train the bias and main model sequentially but found significantly worse results. 

During inference, we discard the bias model predictions $\hat y_b$ and only use $\hat{y}_m$ for unseen events (see \autoref{fig:method:framework}). 
To ensure proper adaptation, we select expert representations corresponding to the given event type, which is known a priori due to the nature of disaster monitoring tasks.

%% file: latex/3_experiments.tex
\begin{table*}[t]
    \centering
    \resizebox{\linewidth}{!}{
    \begin{tabular}{lcccc|cccc|cccc}
        \toprule
        & \multicolumn{4}{c}{\textbf{\textsc{HumAid}}} & \multicolumn{4}{c}{\textbf{\textsc{CrisisLex}}} & \multicolumn{4}{c}{\textbf{\textsc{Trecis}}} \\
        \cmidrule(rl){2-5} \cmidrule(rl){6-9} \cmidrule(rl){10-13}
        \textbf{Method} & \multicolumn{1}{c}{P} & \multicolumn{1}{c}{R} & \multicolumn{1}{c}{F1} & \multicolumn{1}{c}{$\Delta$ (F1)} & \multicolumn{1}{c}{P} & \multicolumn{1}{c}{R} & \multicolumn{1}{c}{F1} & \multicolumn{1}{c}{$\Delta$ (F1)} & \multicolumn{1}{c}{P} & \multicolumn{1}{c}{R} & \multicolumn{1}{c}{F1} & \multicolumn{1}{c}{$\Delta$ (F1)} \\
        \toprule
        \textsc{Bert} & 69.96 & 68.67 & 68.59 & - & 58.99 & 49.58 & 51.84 & - & 56.52 & 49.54 & 50.43 & - \\
        \textsc{Bert-mt} & 68.86 & 69.40 & 68.66 & +0.07 & 51.42 & 50.22 & 49.96 & -1.88 & 51.90 & 47.10 & 47.02 & -3.41 \\
        \midrule
        \multicolumn{1}{l}{\textbf{\textit{Pipeline}}} \\
        \textsc{PoE} & 35.75 & 27.08 & 26.02 & -42.57 & 27.37 & 22.84 & 20.45 & -31.39 & 42.74 & 23.04 & 22.18 & -28.25 \\
        \textsc{Corsair} & 69.39 & 68.85 & 68.50 & -0.09 & 59.71 & 49.15 & 51.53 & -0.31 & 51.48 & 60.12 & 53.03 & +2.60 \\
        \textsc{Masking} & 69.91 & 69.33 & 69.08 & +0.49 & 58.45 & 53.13 & 54.35 & +2.51 & 54.64 & 54.94 & 52.65 & +2.22 \\
        \midrule
        \multicolumn{1}{l}{\textbf{\textit{End2end}}} \\
        \textsc{Nfl-cp} & 68.43 & 67.84 & 67.74 & -0.85 & 55.38 & 49.05 & 50.11 & -1.73 & 63.48 & 42.59 & 42.92 & -7.51 \\
        \textsc{Ear} & 69.46 & 69.22 & 68.66 & +0.07 & 58.23 & 51.83 & 53.48 & +1.64 & 58.22 & 48.78 & 49.81 & -0.62 \\
        \textsc{Eann} & 69.19 & 70.07 & 69.42 & +0.83 & 53.01 & 52.22 & 52.02 & +0.18 & 54.46 & 53.38 & 51.11 & +0.68 \\
        \midrule
        \textsc{Ours} & 70.16 & 71.85 & \textbf{70.71} & \textbf{+2.12\textdagger} & 58.35 & 56.37 & \textbf{56.26} & \textbf{+4.42\textdagger} & 58.08 & 53.96 & \textbf{54.12} & \textbf{+3.69\textdagger} \\
        \bottomrule
    \end{tabular}}
    \caption{Macro Precision, Recall, and F1 scores of the three datasets, averaged and tested over the same five seeds. $\Delta$ represents the difference to \textsc{BERT}. \textbf{Bold} numbers indicate the overall best result, whereas $\dagger$ denotes statistical significance compared to the baselines, except for \textsc{Corsair} on \textsc{Trecis} (paired t-test with p-value < 0.05).}
    \label{table:experiments:main}
\end{table*}

\begin{table}[t]
    \centering
    \resizebox{\linewidth}{!}{
    \begin{tabular}{lcc|cc|cc}
        \toprule
        & \multicolumn{2}{c}{\textbf{\textsc{HumAid}}} & \multicolumn{2}{c}{\textbf{\textsc{CrisisLex}}} & \multicolumn{2}{c}{\textbf{\textsc{Trecis}}} \\
        \cmidrule(rl){2-3} \cmidrule(rl){4-5} \cmidrule(rl){6-7}
        \textbf{Method} & \multicolumn{1}{c}{F1} & \multicolumn{1}{c}{RL} & \multicolumn{1}{c}{F1} & \multicolumn{1}{c}{RL} & \multicolumn{1}{c}{F1} & \multicolumn{1}{c}{RL} \\
        \toprule
        \textsc{Ours} & 70.71 & - & 56.26 & - & 54.12 & - \\
        \midrule
        \textsc{Qwen2.5 1.5B} & 31.70 & 165x & 37.54 & 119x & 21.97 & 131x  \\
        \textsc{Qwen2.5 3B} & 51.75 & 221x & 42.59 & 150x & 22.34 & 156x \\
        \textsc{Qwen2.5 14B} & 57.84 & 409x & 45.27 & 274x & 26.61 & 256x \\
        \textsc{LLaMA3 8B} & 54.23 & 304x & 41.73 & 216x & 23.50 & 194x \\
        \bottomrule
    \end{tabular}}
    \caption{Comparison with instruction-following LLMs. Relative latency (RL) is computed as the ratio of inference time (seconds per document) using a single NVIDIA A100 GPU.}
    \label{table:experiments:llms}
\end{table}

\section{Experiments}\label{sec:experiments}

\paragraph{Datasets}
For our experiments, we use three widely used disaster event classification datasets collected from Twitter, covering various domains (e.g., hurricanes, wildfires, etc) and a diverse range of natural and human-made disasters. \textsc{HumAid} \cite{alam2021humaid} and \textsc{CrisisLex} \cite{olteanu2015crisislex} represent multi-class tasks with a set of important categories for humanitarian aid. Similarly, \textsc{Trecis} \cite{buntain2021trecis} consists of multiple information types but is formulated as a multi-label task.
To simulate the real-world temporal scenario, we use a temporal split strategy and divide the events into disjoint sets according to the provided timestamps.
We provide the dataset details in \ref{appendix:datasets}.

\paragraph{Metrics}
We use macro-averaged precision (P), recall (R), and F1 scores as evaluation metrics to effectively highlight bias since macro F1 is sensitive to skewed performance across all categories. Notably, in the disaster domain, the most critical categories tend to be naturally underrepresented, which supports our choice.

\paragraph{Experimental Setup}
We chose the base-uncased version  of \textsc{Bert} as our PLM and  for all compared baselines but also investigate the use of \textsc{DeBERTa} as strong and robust model in \ref{appendix:ablation:encoder}.
For the bias encoder, we use a CNN composed of five convolutional layers with a channel size of $64$.
The predictors for both the bias and main model consists of a two-layer feedforward network with $384$ hidden units, GELU \citep{hendrycks2016gelu} activation, and is trained with a dropout probability of 20\%.
We set $\alpha=0.1$ and $\lambda=0.2$ for the training stage and additionally mask the bias tokens with a probability of 50\%.
For both training and inference, we select the expert corresponding to the event type (i.e., domain).
In \ref{appendix:details:implementation} and \ref{appendix:details:hyperparameters}, we include more details about training and hyper-parameter selection.

\paragraph{Baselines}
As baselines, we consider the vanilla classifier \textsc{Bert} and a multi-task variant with domain prediction as auxiliary task, called \textsc{Bert-mt}.
Furthermore, we compare our approach with a wide range of debiasing baselines including pipeline and end-to-end (end2end) methods.
The pipeline approaches require bias/spurious tokens while the end2end methods do not neccessiate any further information or only sample-level annotations.
Specifically, for the pipeline methods we include Product-of-Experts (\textsc{PoE}) \citep{clark2019poe}, \textsc{Corsair} \citep{qian2021corsair}, and \textsc{Masking} \citep{wang2022identifying}.
The end2end approaches cover \textsc{Nfl-cp} \citep{chew2024nfl}, \textsc{Ear} \citep{attanasio2022ear}, and the adversarial approach \textsc{Eann} \citep{medina2020eann}.
Additional baseline details are provided in \ref{appendix:baselines}.

\begin{table}[t]
    \centering
    \resizebox{\linewidth}{!}{
    \begin{tabular}{lcc|cc|cc}
        \toprule
        & \multicolumn{2}{c}{\textbf{\textsc{HumAid}}} & \multicolumn{2}{c}{\textbf{\textsc{CrisisLex}}} & \multicolumn{2}{c}{\textbf{\textsc{Trecis}}} \\
        \cmidrule(rl){2-3} \cmidrule(rl){4-5} \cmidrule(rl){6-7}
        \textbf{Method} & \multicolumn{1}{c}{F1} & \multicolumn{1}{c}{$\Delta$ (F1)} & \multicolumn{1}{c}{F1} & \multicolumn{1}{c}{$\Delta$ (F1)} & \multicolumn{1}{c}{F1} & \multicolumn{1}{c}{$\Delta$ (F1)} \\
        \toprule
        \textsc{Bert} & 68.59 & -  & 51.84 & -  & 50.43 & - \\
        \midrule
        \textsc{Ours} & 70.71 & +2.12 & 56.26 & +4.42 & 54.12 & +3.69 \\
        1\hspace{0.2em} w/o experts & 70.54 & +1.95 & 54.60 & +2.76 & 53.32 & +2.89 \\
        2\hspace{0.2em} w/o debias & 70.17 & +1.58  & 54.55 & +2.71 & 53.76 & +3.33 \\
        3\hspace{0.2em} w/o augment & 70.29 & +1.70 & 55.49 & +3.65 & 53.91 & +3.48 \\
        \bottomrule
    \end{tabular}}
    \caption{Ablation results for macro F1 scores.}
    \label{table:experiments:ablation}
\end{table}

\subsection{Results}\label{sec:experiments:results}

In \autoref{table:experiments:main}, we present macro P, R, and F1 scores for the main experiments. 
Our approach consistently outperforms all baselines across the three tasks in terms of F1 score, achieving improvements of at least +1.3\% on \textsc{HumAid}, +1.9\% on \textsc{CrisisLex}, and + 1.1\% on \textsc{Trecis}, respectively.
These results confirm the effectiveness of our method.
Surprisingly, the second-best method is the simple \textsc{Masking} approach, which underscores the impact of event-related bias to unseen disaster events.

Notably, our method surpasses \textsc{Eann} -- the closest work -- by up to +4.2\% in P, R, and F1. 
The authors highlight in their work that \textsc{Eann} struggles with mixed event types when using adversarial bias removal.
This necessitates training a separate model for each event type.
Our experts overcome these limitations by disentangling disaster domains.

We also explored training the bias model separately (\textsc{PoE}), as suggested by \citet{clark2019poe}, rather than jointly with the main model. 
However, this consistently led to poor performance, likely because the bias model captures more than just bias, hindering the main model’s ability to learn essential features (see \autoref{fig:experiments:probing}).

\subsection{Comparison with LLMs}\label{sec:experiments:llms}

In addition to PLM-based baselines, we compare our approach to instruction-following LLMs which offer an alternative without requiring task-specific fine-tuning.
We evaluate the Qwen2.5 \cite{qwen2025} model series and LLaMA3-8B \cite{llama2024} for comparison.
As prompt, we use the annotation guidelines of the datasets and provide one example for each information type.
\autoref{table:experiments:llms} reports the macro F1 scores and the relative latency (RL).
Although LLMs achieve promising results, their substantially higher inference latency may limit their practicality for real-time or large-scale disaster response applications.

\subsection{Analysis}\label{sec:experiments:analysis}

\paragraph{Ablation Study} 
To illustrate the effectiveness of the proposed components, we conduct ablation studies and present the results in \autoref{table:experiments:ablation}. 
In line 1, we assess the impact of our experts by replacing it with a unified query mechanism, which leads to a performance drop of up to -1.7\% in F1, while \textsc{HumAid} experiences only a slight decrease of -0.2\%.
In line 2, removing the bias model results in the largest performance drop (up to -1.7\%), highlighting the importance of event-related bias removal. 
Lastly, in line 3, we remove masking augmentation, revealing its complementary nature to our approach, as it constantly improves performance by only masking spurious tokens.

\paragraph{Probing Design} 
Since analyzing all samples is infeasible, we design a probing test (see \ref{appendix:probing}) to evaluate the information encoded by both the bias model and the main model components. 
To achieve this, we encode the training set for \textsc{HumAid} and \textsc{CrisisLex} using the baseline, bias, and main model. 
We then fit a shallow classifier using only a small subset of the samples (5\%) to assess the representations. 
The probing tasks include predicting domains, events, and information types. 
To ensure robustness, we run the probing tasks with 25 seeds and report the averaged macro F1 scores.

\begin{figure}[t]
    \centering
    \includegraphics[width=1.0\linewidth]{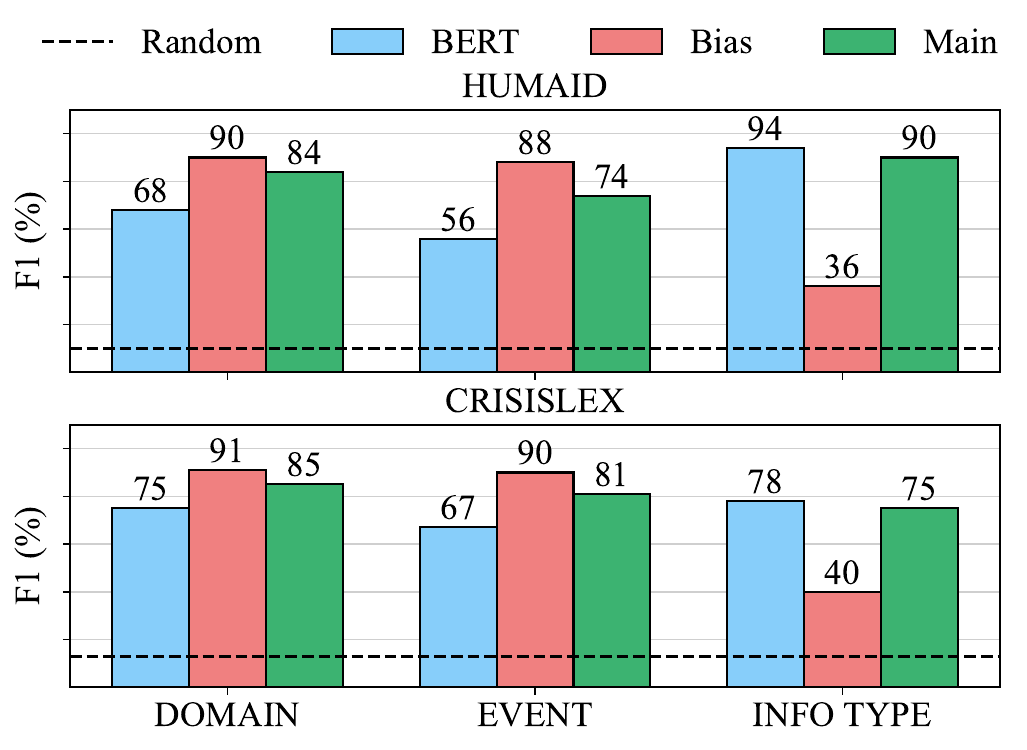}
    \caption{Macro F1 scores (averaged over 25 runs) for the designed probing tasks and our focused models. The random baseline refers to the information type classification task.}
    \label{fig:experiments:probing}
\end{figure}

\paragraph{Probing Insights} 
As shown in \autoref{fig:experiments:probing}, the bias and main components exhibit opposing behaviors: the event-related bias model achieves high F1 scores for domains and events but suffers a significant performance drop in the main task, whereas the debiased main model excels in the main task.
This supports our design decisions to mitigate overfitting of event-related bias.
Interestingly, the baseline model achieves the highest main task performance for train events but worse performance for test events (see \autoref{table:experiments:main}), supporting our hypothesis that it assigns greater weight to the seen event-specific features, while not retaining more general features.
In contrast, our model's high domain scores can be attributed to the experts, which facilitate domain-separated attention distributions and representations.

%% file: latex/4_conclusion.tex
\section{Conclusion}\label{sec:conclusion}

In this work, we explore event- and domain-related biases through the lens of causality and propose a method to enhance the generalization for unseen disaster events.
Specifically, we design a bias model to mitigate the influence of event-related tokens, such as locations, retrieval keywords, and numbers. 
Additionally, we introduce an expert-based component to reduce domain bias caused by overrepresented event types. 
Our experiments on three challenging disaster event datasets demonstrate the effectiveness of our approach.
For future work, we aim to address additional bias factors, including temporal shifts (e.g., initial vs. recovery phases), regional disparities (e.g., marginalized vs. politically significant areas), and LLM-based data augmentation to simulate communication patterns across a wider range of disaster events.

%% file: latex/5_limitations.tex
\section*{Limitations}\label{sec:limitations}
Our study focuses solely on disaster event classification, making pre-trained NER systems and simple heuristics suitable for detecting most event-related bias features. 
However, these systems are prone to errors, and the reliance on this types of features may limit their applicability to other use cases. 
Therefore, efficient automatic methods must be explored to bypass the need for tailored bias feature identification \citep{wang2022identifying}.
Although social networks are a valuable supplement to formal sources (e.g., news), our considered datasets consists only of Twitter data in English language, which includes platform-specific features such as hashtags. 
Expanding the scope to other data sources, event types, and languages would lead to better assessment of generalizability \citep{mccreadie2023crisisfacts}. 
However, the lack of annotated data remains a significant challenge. 
A large and diverse corpus covering multiple events, sources, and languages could help to mitigate bias effects during the offline training phase.

%% file: latex/6_ethics.tex
\section*{Ethical Considerations}\label{sec:limitations}

Automated disaster classification systems can inadvertently harm vulnerable populations if errors misdirect aid or overlook critical needs.
For instance, false negatives could delay relief efforts in areas that require urgent support.
Social media data may also reflect societal biases, platforms such as X (formerly Twitter) are unevenly available across regions, and not all individuals have access to these platforms.
This can propagate inequities in response efforts. 
While our method takes a step toward reducing event- and domain-related biases, careful evaluation and transparent reporting remain essential for responsible deployment.

%% file: latex/7_appendix.tex
\subsection{Datasets}\label{appendix:datasets}

We experiment with three publicly available Twitter datasets that support both multi-class and multi-label information type classification. 
These datasets were selected for their large number of events and diverse range of event types (i.e., domains), including natural and human-made disasters. 
To simulate real-world scenarios, we apply a temporal split strategy, dividing the events into disjoint training, validation, and test sets based on their timestamps.
Detailed dataset statistics are provided in \autoref{table:appendix:datasets}.

\paragraph{\textsc{HumAid}} 
This dataset is composed of tweets from 19 disaster-related events, with sizes ranging from approximately 570 to 9,500 tweets \cite{alam2021humaid}. 
\textsc{HumAID} categorizes posts into 11 classes, spanning from \textit{Not Humanitarian} to \textit{Injured or Dead People} and provides fine-grained insights into ongoing disasters. 
To maintain consistency across temporal splits, we exclude posts labeled \textit{Can't Judge} and merge the closely related categories \textit{Injured or Dead People} and \textit{Missing or Found People}.

\paragraph{\textsc{CrisisLex}} 
The \textsc{T26} variant of \textsc{CrisisLex} \cite{olteanu2015crisislex} contains annotated tweets from 26 crisis events, formulated as multi-class task with seven information types including the category \textit{Not Related}.
This dataset captures a diverse set of emergency events with approximately 1,000 tweets per individual event.
As a preprocessing step, we removed tweets labeled \textit{Not Labeled} to ensure data quality.

\paragraph{\textsc{Trecis}} 
TREC Incident Streams is a multi-label classification task comprising over 70 events and annotations for 25 information types \cite{buntain2021trecis}.
The dataset varies significantly in size, ranging from 90 to 5,900 tweets per event, with diverse label distributions.
For our experiments, we exclude \textit{COVID} events and retain only those with at least 400 tweets and multiple relevant labels, as some events primarily contain irrelevant information types.
Additionally, we hierarchically merge semantically similar categories to address the issue of extremely low-frequency labels.

\begin{table*}[t]
    \centering
    \vspace{-2mm}
    \resizebox{\linewidth}{!}{
    \begin{tabular}{llllccc}
        \toprule
        \textbf{Dataset} & \textbf{Domains} & \textbf{Events} & \textbf{Labels} & \textbf{Train} & \textbf{Valid} & \textbf{Test} \\
        \toprule
        \textsc{HumAid} & earthquake, hurricane, floods, wildfire & 19 (12, 3, 4) & injured or dead people & 3963 & 2785 & 913 \\
         &  &  & rescue, volunteering or donation & 10295 & 7187 & 3796 \\
         &  &  & sympathy and support & 4599 & 1779 & 2553  \\
         &  &  & infrastructure and utility damage & 5691 & 1038 & 1434 \\
         &  &  & requests or urgent needs & 1508 & 723 & 387 \\
         &  &  & caution and advice & 2198 & 1588 & 1608 \\
         &  &  & displaced people and evacuations & 2079 & 1061 & 859 \\
         &  &  & other relevant information & 7237 & 2629 & 2278 \\
         &  &  & not humanitarian & 1906 & 2834 & 1556 \\
         & & & \textbf{all} & 39476 & 21624 & 15384 \\
         \midrule
        \textsc{CrisisLex} & earthquake, typhoon, floods, wildfire,  & 26 (15, 5, 6) & affected individuals & 2339 & 714 & 1206 \\
         & accident, other &  & donations and volunteering & 1081 & 500 & 519 \\
         &  &  & sympathy and support & 2173 & 577 & 1168  \\
         &  &  & infrastructure and utilities & 850 & 326 & 285 \\
         &  &  & caution and advice & 1277 & 493 & 249 \\
         &  &  & other relevant information & 4104 & 1475 & 1265 \\
         &  &  & not relevant & 752 & 167 & 102 \\
         & & & \textbf{all} & 12576 & 4252 & 4794 \\
         \midrule
        \textsc{Trecis} & earthquake, hurricane, floods, wildfire,  & 33 (16, 8, 9) & observations and facts & 10542 & 4439 & 6259 \\
         & accident &  & rescue, volunteering or donation & 984 & 223 & 306 \\
         &  &  & sentiment, sympathy or discussion & 9966 & 1196 & 922 \\
         &  &  & requests or urgent needs & 531 & 63 & 74 \\
         &  &  & caution, news or reports & 8456 & 3126 & 3485 \\
         &  &  & other relevant information & 4515 & 5472 & 7234 \\
         &  &  & not relevant & 4582 & 2690 & 1983 \\
         & & & \textbf{all} & 25221 & 8688 & 9365 \\
        \bottomrule
    \end{tabular}}
    \caption{Detailed statistics of the three datasets. The number in the brackets corresponds to the number of events in train, validation, and test sets.}
    \label{table:appendix:datasets}
    \vspace{-2mm}
\end{table*}

\begin{figure*}[t]
  \centering
\includegraphics[width=1.0\linewidth]{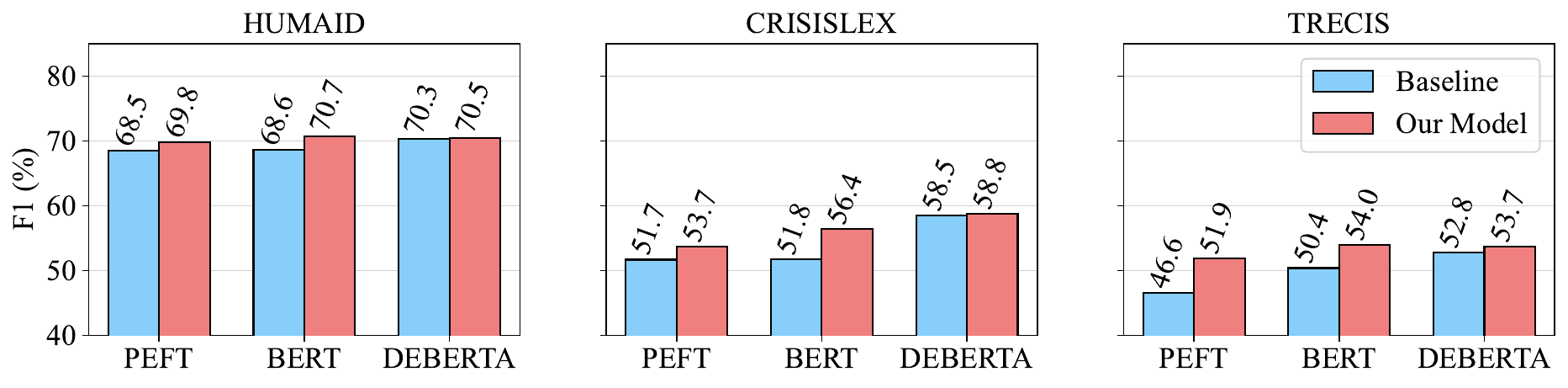}
  \caption{Macro F1 scores for different PLM encoders. Baseline represents a simple classification head. The results are the average of five runs.}
  \label{fig:ablation:encoders}
  \vspace{-2mm}
\end{figure*}

\subsection{Baselines}\label{appendix:baselines}

We consider a diverse set of debiasing baselines, encompassing both pipeline and end-to-end methods. 
Pipeline approaches rely on explicitly identified bias features and tokens, whereas end2end methods operate without requiring such information or, at most, depend on sample-level annotations.
The pipeline approaches include \textsc{PoE}, \textsc{Corsair}, and \textsc{Masking}.
For end-to-end methods, we adopt \textsc{NFL-CP}, \textsc{EAR}, and \textsc{EANN}.
Each of these approaches is described in more detail below.

\paragraph{\textsc{PoE}} Product-of-Experts \citep{clark2019poe} is widely used in many existing works in order to remove known dataset biases.
These works first train a bias model to capture the known biases and then train the main model as ensemble together with the bias model predictions.
After training, only the main model will be used for inference.
We train the bias model with the event-related tokens as input.

\paragraph{\textsc{Corsair}} first trains a biased model on the training set directly and then applies counterfactual inference on this biased model \citep{qian2021corsair}.
Their method generates two types of counterfactual documents to produce outputs to distill the label and keyword bias.
During inference, the distilled biases are removed from the original predictions.
We use the identified event-related tokens as context words.

\paragraph{\textsc{Masking}} represents a simple but effective augmentation strategy which replaces tokens or spans with the \textsc{BERT}-specific [MASK] token.
In this work, we follow \citet{wang2022identifying} and mask the identified event-related tokens in the training phase with a probability of 80\%.
In this way, we aim to motivate the model to focus on context words and to learn the general patterns.

\paragraph{\textsc{Nfl-cp}} This model family mitigates spurious correlations with regularization techniques that aim to prevent token misalignments \citep{chew2024nfl}.
We use the Constrained Parameters variant which penalizes large changes in the PLM parameters.
For our experiments, we rely on the regularization hyper-parameter $\lambda=15000$ as proposed by the authors.

\paragraph{\textsc{Ear}} Entropy-based Attention Regularization represents a knowledge-free bias mitigation method without the need of any known spurious terms \citep{attanasio2022ear}.
The authors penalize tokens with low attention entropy and therefore low contextualization.
We follow the authors original setting and compute the regularization loss for each transformer layer and set the regularization strength as $\alpha=0.01$.

\paragraph{\textsc{Eann}} is the most related work to the focused disaster event domain.
This approach adds an additional adversarial model component designed for event-related bias removal \citep{medina2020eann}.
We follow the proposed framework and implement the adversarial model with a gradient-reversal layer and scaling parameter $\lambda=1$.
For the multi-label task (\textsc{Trecis}), we set the adversarial weight to $0.2$ in order to achieve more stable results.
We train a joint model across all event types to enable a fair comparison.

\subsection{Implementation Details}\label{appendix:details:implementation}

In our experiments, we use the implementation of the \textit{Transformers} \citep{wolf2020transformers} (v4.47.0) and \textit{Lightning}\footnote{https://lightning.ai/docs/pytorch/stable/} (v2.4.0) library in conjunction with PyTorch (v2.3.0).
For all runs, we apply five seeds (0, 42, 64, 86, 128) and report the average scores as results.
All models are trained for 30 epochs with Adam \citep{kingma2014} and cosine scheduling without warmup steps but experienced no improvement after 10 epochs for \textsc{HumAid} and \textsc{CrisisLex}, respectively.
We evaluated the performance at each epoch on the validation set and selected the best-performing checkpoint.
Unless otherwise mentioned, we use bert-base-uncased\footnote{google-bert/bert-base-uncased} \citep{devlin2019bert} with 110M parameters as PLM.
For NER we use a model\footnote{TweebankNLP/bertweet-tb2\_wnut17-ner} specifically trained for Twitter.
All models are trained with A100 GPUs with 40GB HBM using compute nodes running CUDA 12.3.

\subsection{Hyper-parameter Tuning}\label{appendix:details:hyperparameters}

We tune the hyper-parameters learning rate ($1 \times 10^{-5}$, $3 \times 10^{-5}$, $5 \times 10^{-5}$, $1 \times 10^{-4}$), batch size ($8$, $16$, $32$, $64$), $\lambda$ (0.2, 0.5, 1.0), and $\alpha$ (0.1, 0.2, 0.3) for the first seed and validation sets. 
The models performed best using the batch size $32$ and learning rate $1 \times 10^{-5}$ for the encoder and $1 \times 10^{-4}$ for the remaining initialized parameters.
In addition, we found with $\lambda=0.2$ and $\alpha=0.1$ the most robust setup.
Due to computational constraints, we apply the best setting of the first seed to the remaining runs.

\subsection{Encoder Ablation}\label{appendix:ablation:encoder}

With \textsc{DeBERTa}, we test a more robustly trained PLM that has demonstrated superior results for a variety of tasks.
Additionally, we apply our method with Parameter-Efficient Fine-Tuning (\textsc{PEFT}) using AdaLoRA \citep{zhang2023adalora} in conjunction with \textsc{BERT}.
The comparison of \textsc{PEFT}, \textsc{BERT}, and \textsc{DeBERTa}\footnote{microsoft/deberta-v3-base} is shown in \autoref{fig:ablation:encoders}.
Our results show consistent improvements with \textsc{PEFT} and slight improvements with \textsc{DeBERTa} across all three datasets. 
In particular, the robustness gap narrows when using more robust models with better initial representations.
This aligns with findings from other research \cite{chew2024nfl}.

\subsection{Probing Details}\label{appendix:probing}

To analyze the model's behavior, we design probing tasks to evaluate the information encoded by the baseline, bias, and main models. Specifically, we leverage existing annotations to introduce tasks for predicting the domains, events, and information types (i.e., the main task). 
To assess the encoded information, we encode the training sets for \textsc{HumAid} and \textsc{CrisisLex} using the relevant model components. 
We then train a shallow classifier using only a small subset (5\%) of the training samples to evaluate the learned representations. The rationale behind this setup is to examine potential overfitting to event-related bias associated with previously seen events. 
For the shallow model, we use a logistic regression classifier without any adaptations or preprocessing applied to the encodings. To ensure the robustness of our results, we perform the probing tasks across 25 different random seeds and train-test splits.
Finally, we report the average scores from these runs.

%% file: latex/acl_latex.bbl
\begin{thebibliography}{29}
\providecommand{\natexlab}[1]{#1}

\bibitem[{Alam et~al.(2018)Alam, Joty, and Imran}]{alam2018domainadaptation}
Firoj Alam, Shafiq Joty, and Muhammad Imran. 2018.
\newblock \href {https://doi.org/10.18653/v1/P18-1099} {Domain adaptation with adversarial training and graph embeddings}.
\newblock In \emph{Proceedings of the 56th Annual Meeting of the Association for Computational Linguistics (Volume 1: Long Papers)}, pages 1077--1087, Melbourne, Australia. Association for Computational Linguistics.

\bibitem[{Alam et~al.(2021)Alam, Qazi, Imran, and Ofli}]{alam2021humaid}
Firoj Alam, Umair Qazi, Muhammad Imran, and Ferda Ofli. 2021.
\newblock \href {https://doi.org/10.1609/icwsm.v15i1.18116} {Humaid: Human-annotated disaster incidents data from twitter with deep learning benchmarks}.
\newblock \emph{Proceedings of the International AAAI Conference on Web and Social Media}, 15(1):933--942.

\bibitem[{Attanasio et~al.(2022)Attanasio, Nozza, Hovy, and Baralis}]{attanasio2022ear}
Giuseppe Attanasio, Debora Nozza, Dirk Hovy, and Elena Baralis. 2022.
\newblock \href {https://doi.org/10.18653/v1/2022.findings-acl.88} {Entropy-based attention regularization frees unintended bias mitigation from lists}.
\newblock In \emph{Findings of the Association for Computational Linguistics: ACL 2022}, pages 1105--1119, Dublin, Ireland. Association for Computational Linguistics.

\bibitem[{Buntain et~al.(2021)Buntain, McCreadie, and Soboroff}]{buntain2021trecis}
Cody~L. Buntain, Richard McCreadie, and Ian Soboroff. 2021.
\newblock Incident {Streams} 2020: {TREC}-{IS} in the {Time} of {COVID}-19.
\newblock In \emph{ISCRAM 2021: 18th International Conference on Information Systems for Crisis Response and Management}.

\bibitem[{Chen et~al.(2023)Chen, Hu, Li, Shao, and Nie}]{chen2023causal}
Ziwei Chen, Linmei Hu, Weixin Li, Yingxia Shao, and Liqiang Nie. 2023.
\newblock \href {https://doi.org/10.18653/v1/2023.acl-long.37} {Causal intervention and counterfactual reasoning for multi-modal fake news detection}.
\newblock In \emph{Proceedings of the 61st Annual Meeting of the Association for Computational Linguistics (Volume 1: Long Papers)}, pages 627--638, Toronto, Canada. Association for Computational Linguistics.

\bibitem[{Chew et~al.(2024)Chew, Lin, Chang, and Huang}]{chew2024nfl}
Oscar Chew, Hsuan-Tien Lin, Kai-Wei Chang, and Kuan-Hao Huang. 2024.
\newblock \href {https://aclanthology.org/2024.findings-eacl.68/} {Understanding and mitigating spurious correlations in text classification with neighborhood analysis}.
\newblock In \emph{Findings of the Association for Computational Linguistics: EACL 2024}, pages 1013--1025, St. Julian{'}s, Malta. Association for Computational Linguistics.

\bibitem[{Clark et~al.(2019)Clark, Yatskar, and Zettlemoyer}]{clark2019poe}
Christopher Clark, Mark Yatskar, and Luke Zettlemoyer. 2019.
\newblock \href {https://doi.org/10.18653/v1/D19-1418} {Don`t take the easy way out: Ensemble based methods for avoiding known dataset biases}.
\newblock In \emph{Proceedings of the 2019 Conference on Empirical Methods in Natural Language Processing and the 9th International Joint Conference on Natural Language Processing (EMNLP-IJCNLP)}, pages 4069--4082, Hong Kong, China. Association for Computational Linguistics.

\bibitem[{Devlin et~al.(2019)Devlin, Chang, Lee, and Toutanova}]{devlin2019bert}
Jacob Devlin, Ming-Wei Chang, Kenton Lee, and Kristina Toutanova. 2019.
\newblock \href {https://doi.org/10.18653/v1/N19-1423} {{BERT}: Pre-training of deep bidirectional transformers for language understanding}.
\newblock In \emph{Proceedings of the 2019 Conference of the North {A}merican Chapter of the Association for Computational Linguistics: Human Language Technologies, Volume 1 (Long and Short Papers)}, pages 4171--4186, Minneapolis, Minnesota. Association for Computational Linguistics.

\bibitem[{Grattafiori et~al.(2024)}]{llama2024}
Aaron Grattafiori et~al. 2024.
\newblock \href {https://arxiv.org/abs/2407.21783} {The llama 3 herd of models}.
\newblock \emph{Preprint}, arXiv:2407.21783.

\bibitem[{Hendrycks and Gimpel(2016)}]{hendrycks2016gelu}
Dan Hendrycks and Kevin Gimpel. 2016.
\newblock \href {https://doi.org/10.48550/ARXIV.1606.08415} {Gaussian {Error} {Linear} {Units} ({GELUs})}.
\newblock \emph{arXiv preprint}.
\newblock Version Number: 5.

\bibitem[{Kaufhold(2021)}]{kaufhold2021crisis}
Marc-André Kaufhold. 2021.
\newblock \href {https://doi.org/10.1007/978-3-658-33341-6} {\emph{Information {Refinement} {Technologies} for {Crisis} {Informatics}: {User} {Expectations} and {Design} {Principles} for {Social} {Media} and {Mobile} {Apps}}}.
\newblock Springer Fachmedien Wiesbaden, Wiesbaden.

\bibitem[{Kingma and Ba(2014)}]{kingma2014}
Diederik Kingma and Jimmy Ba. 2014.
\newblock Adam: A method for stochastic optimization.
\newblock \emph{International Conference on Learning Representations}.

\bibitem[{Kruspe et~al.(2021)Kruspe, Kersten, and Klan}]{kruspe2021review}
Anna Kruspe, Jens Kersten, and Friederike Klan. 2021.
\newblock \href {https://doi.org/10.5194/nhess-21-1825-2021} {Review article: Detection of actionable tweets in crisis events}.
\newblock \emph{Natural Hazards and Earth System Sciences}, 21(6):1825--1845.

\bibitem[{McCreadie and Buntain(2023)}]{mccreadie2023crisisfacts}
Richard McCreadie and Cody~L. Buntain. 2023.
\newblock Crisisfacts: Buidling and evaluating crisis timelines.
\newblock In \emph{ISCRAM 2023: 20th International Conference on Information Systems for Crisis Response and Management}.

\bibitem[{Medina~Maza et~al.(2020)Medina~Maza, Spiliopoulou, Hovy, and Hauptmann}]{medina2020eann}
Salvador Medina~Maza, Evangelia Spiliopoulou, Eduard Hovy, and Alexander Hauptmann. 2020.
\newblock \href {https://doi.org/10.18653/v1/2020.findings-emnlp.344} {Event-related bias removal for real-time disaster events}.
\newblock In \emph{Findings of the Association for Computational Linguistics: EMNLP 2020}, pages 3858--3868, Online. Association for Computational Linguistics.

\bibitem[{Olteanu et~al.(2015)Olteanu, Vieweg, and Castillo}]{olteanu2015crisislex}
Alexandra Olteanu, Sarah Vieweg, and Carlos Castillo. 2015.
\newblock \href {https://doi.org/10.1145/2675133.2675242} {What to expect when the unexpected happens: Social media communications across crises}.
\newblock In \emph{Proceedings of the 18th ACM Conference on Computer Supported Cooperative Work \& Social Computing}, CSCW '15, page 994–1009, New York, NY, USA. Association for Computing Machinery.

\bibitem[{Qian et~al.(2021)Qian, Feng, Wen, Ma, and Xie}]{qian2021corsair}
Chen Qian, Fuli Feng, Lijie Wen, Chunping Ma, and Pengjun Xie. 2021.
\newblock \href {https://doi.org/10.18653/v1/2021.acl-long.422} {Counterfactual inference for text classification debiasing}.
\newblock In \emph{Proceedings of the 59th Annual Meeting of the Association for Computational Linguistics and the 11th International Joint Conference on Natural Language Processing (Volume 1: Long Papers)}, pages 5434--5445, Online. Association for Computational Linguistics.

\bibitem[{Reuter et~al.(2018)Reuter, Hughes, and Kaufhold}]{reuter2018crisis}
Christian Reuter, Amanda~Lee Hughes, and Marc-André Kaufhold. 2018.
\newblock \href {https://doi.org/10.1080/10447318.2018.1427832} {Social {Media} in {Crisis} {Management}: {An} {Evaluation} and {Analysis} of {Crisis} {Informatics} {Research}}.
\newblock \emph{International Journal of Human–Computer Interaction}, 34(4):280--294.

\bibitem[{Sakaki et~al.(2010)Sakaki, Okazaki, and Matsuo}]{sakaki2010earthquake}
Takeshi Sakaki, Makoto Okazaki, and Yutaka Matsuo. 2010.
\newblock \href {https://doi.org/10.1145/1772690.1772777} {Earthquake shakes twitter users: Real-time event detection by social sensors}.
\newblock In \emph{Proceedings of the 19th International Conference on World Wide Web}, WWW '10, page 851–860, New York, NY, USA. Association for Computing Machinery.

\bibitem[{Seeberger and Riedhammer(2022)}]{seeberger2022emlm}
Philipp Seeberger and Korbinian Riedhammer. 2022.
\newblock \href {https://aclanthology.org/2022.nlp4pi-1.9} {Enhancing crisis-related tweet classification with entity-masked language modeling and multi-task learning}.
\newblock In \emph{Proceedings of the Second Workshop on NLP for Positive Impact (NLP4PI)}, pages 70--78, Abu Dhabi, United Arab Emirates (Hybrid). Association for Computational Linguistics.

\bibitem[{Wang et~al.(2022)Wang, Sridhar, Yang, and Wang}]{wang2022identifying}
Tianlu Wang, Rohit Sridhar, Diyi Yang, and Xuezhi Wang. 2022.
\newblock \href {https://doi.org/10.18653/v1/2022.findings-naacl.130} {Identifying and mitigating spurious correlations for improving robustness in {NLP} models}.
\newblock In \emph{Findings of the Association for Computational Linguistics: NAACL 2022}, pages 1719--1729, Seattle, United States. Association for Computational Linguistics.

\bibitem[{Wei et~al.(2021)Wei, Feng, Chen, Wu, Yi, and He}]{wei2021macr}
Tianxin Wei, Fuli Feng, Jiawei Chen, Ziwei Wu, Jinfeng Yi, and Xiangnan He. 2021.
\newblock \href {https://doi.org/10.1145/3447548.3467289} {Model-agnostic counterfactual reasoning for eliminating popularity bias in recommender system}.
\newblock In \emph{Proceedings of the 27th ACM SIGKDD Conference on Knowledge Discovery \& Data Mining}, KDD '21, page 1791–1800, New York, NY, USA. Association for Computing Machinery.

\bibitem[{Wiegmann et~al.(2020)Wiegmann, Kersten, Klan, Potthast, and Stein}]{wiegmann2020analysis}
Matti Wiegmann, Jens Kersten, Friederike Klan, Martin Potthast, and Benno Stein. 2020.
\newblock \href {https://doi.org/10.5281/ZENODO.3713920} {Analysis of {Detection} {Models} for {Disaster}-{Related} {Tweets}}.
\newblock In \emph{ISCRAM 2020: 17th International Conference on Information Systems for Crisis Response and Management}.

\bibitem[{Wolf et~al.(2020)Wolf, Debut, Sanh, Chaumond, Delangue, Moi, Cistac, Rault, Louf, Funtowicz, Davison, Shleifer, von Platen, Ma, Jernite, Plu, Xu, Le~Scao, Gugger, Drame, Lhoest, and Rush}]{wolf2020transformers}
Thomas Wolf, Lysandre Debut, Victor Sanh, Julien Chaumond, Clement Delangue, Anthony Moi, Pierric Cistac, Tim Rault, Remi Louf, Morgan Funtowicz, Joe Davison, Sam Shleifer, Patrick von Platen, Clara Ma, Yacine Jernite, Julien Plu, Canwen Xu, Teven Le~Scao, Sylvain Gugger, Mariama Drame, Quentin Lhoest, and Alexander Rush. 2020.
\newblock \href {https://doi.org/10.18653/v1/2020.emnlp-demos.6} {Transformers: State-of-the-art natural language processing}.
\newblock In \emph{Proceedings of the 2020 Conference on Empirical Methods in Natural Language Processing: System Demonstrations}, pages 38--45, Online. Association for Computational Linguistics.

\bibitem[{Wu et~al.(2024)Wu, Liu, Zhao, Lu, Zhang, Sun, Wu, and Kuang}]{wu2024das}
Yiquan Wu, Yifei Liu, Ziyu Zhao, Weiming Lu, Yating Zhang, Changlong Sun, Fei Wu, and Kun Kuang. 2024.
\newblock \href {https://doi.org/10.1609/aaai.v38i17.29897} {De-biased attention supervision for text classification with causality}.
\newblock \emph{Proceedings of the AAAI Conference on Artificial Intelligence}, 38(17):19279--19287.

\bibitem[{Yang et~al.(2025)}]{qwen2025}
An~Yang et~al. 2025.
\newblock \href {https://arxiv.org/abs/2412.15115} {Qwen2.5 technical report}.
\newblock \emph{Preprint}, arXiv:2412.15115.

\bibitem[{Zhang et~al.(2024)Zhang, Li, Liu, Wu, Wang, and Wang}]{zhan2024evolving}
Jiajun Zhang, Zhixun Li, Qiang Liu, Shu Wu, Zilei Wang, and Liang Wang. 2024.
\newblock \href {https://doi.org/10.1145/3627673.3679919} {Evolving to the future: Unseen event adaptive fake news detection on social media}.
\newblock In \emph{Proceedings of the 33rd ACM International Conference on Information and Knowledge Management}, CIKM '24, page 4273–4277, New York, NY, USA. Association for Computing Machinery.

\bibitem[{Zhang et~al.(2023)Zhang, Chen, Bukharin, He, Cheng, Chen, and Zhao}]{zhang2023adalora}
Qingru Zhang, Minshuo Chen, Alexander Bukharin, Pengcheng He, Yu~Cheng, Weizhu Chen, and Tuo Zhao. 2023.
\newblock \href {https://openreview.net/forum?id=lq62uWRJjiY} {Adaptive budget allocation for parameter-efficient fine-tuning}.
\newblock In \emph{The Eleventh International Conference on Learning Representations}.

\bibitem[{Zhu et~al.(2022)Zhu, Sheng, Cao, Li, Wang, and Zhuang}]{zhu2022endef}
Yongchun Zhu, Qiang Sheng, Juan Cao, Shuokai Li, Danding Wang, and Fuzhen Zhuang. 2022.
\newblock \href {https://doi.org/10.1145/3477495.3531816} {Generalizing to the future: Mitigating entity bias in fake news detection}.
\newblock In \emph{Proceedings of the 45th International ACM SIGIR Conference on Research and Development in Information Retrieval}, SIGIR '22, page 2120–2125, New York, NY, USA. Association for Computing Machinery.

\end{thebibliography}
